# An axiomatic approach to the roughness measure of rough sets

Ping Zhu[a,b]

[a] School of Science, Beijing University of Posts and Telecommunications, Beijing 100876, China
[b] State Key Laboratory of Networking and Switching, Beijing University of Posts and Telecommunications, Beijing 100876, China

**Abstract**

In Pawlak's rough set theory, a set is approximated by a pair of lower and upper approximations. To measure numerically the roughness of an approximation, Pawlak introduced a quantitative measure of roughness by using the ratio of the cardinalities of the lower and upper approximations. Although the roughness measure is effective, it has the drawback of not being strictly monotonic with respect to the standard ordering on partitions. Recently, some improvements have been made by taking into account the granularity of partitions. In this paper, we approach the roughness measure in an axiomatic way. After axiomatically defining roughness measure and partition measure, we provide a unified construction of roughness measure, called strong Pawlak roughness measure, and then explore the properties of this measure. We show that the improved roughness measures in the literature are special instances of our strong Pawlak roughness measure and introduce three more strong Pawlak roughness measures as well. The advantage of our axiomatic approach is that some properties of a roughness measure follow immediately as soon as the measure satisfies the relevant axiomatic definition.

## 1. Introduction

Rough set theory was proposed by Pawlak in the early 1980s [22, 23] as a new mathematical approach for dealing with inexact, uncertain or vague knowledge in information systems. Since then we have witnessed a systematic, world-wide growth of interest in rough set theory [4, 20, 25, 28, 37, 43, 44, 48, 49, 50, 52]. Nowadays, it is widely recognized that rough set applications have a great importance in several fields, such as granular computing, data mining, and approximate reasoning [19, 26, 27, 46, 51, 53].

A basic hypothesis in rough set theory is that some elements of a universe may be indiscernible in view of the available information about the elements. Such an indiscernibility relation was first described by equivalence relation in the way that two elements are related by the relation if and only if they are indiscernible from each other [22, 23]. In this framework, a rough set is a formal approximation of a subset of the universe in terms of a pair of unions of equivalence classes which give the lower and upper approximations of the subset. In order to measure numerically the accuracy of an approximation, Pawlak introduced two quantitative measures of accuracy and roughness in [23]: The accuracy of a subset is defined as the ratio of the cardinalities of the lower and upper approximations of the subset, which is bounded by 0 and 1; the roughness of a subset is calculated by subtracting the accuracy of the subset from 1. Therefore, roughness is a complementary concept to the accuracy of approximation. The roughness is, in some sense, the amount of uncertainty of the underlying subset. A roughness of 1 shows that we have no certain knowledge on the underlying subset, and a roughness of 0 means we know everything for sure about the underlying subset. From this point of view, the roughness measure is an important indicator of the uncertainty and accuracy associated with a given subset.

It has been observed [2, 17, 35] that the roughness (and its dual, accuracy) due to Pawlak, however, has the drawback of not being strictly monotonic with respect to the standard ordering on partitions. In other words, Pawlak's roughness and accuracy measures do not necessarily provide us with information on the uncertainty related to the granularity of partitions. To modify such measures, Beaubouef, Petry, and Arora proposed the notion of rough entropy in [2]. This limitation has also been improved by Xu, Zhou, and Lu in [35] by using the so-called excess entropy approach. Most recently, Liang, Wang, and Qian made another improvement by exploiting the notion of knowledge granulation in [17]. All of these improvements share some good properties. Nevertheless, there exists no unified description for roughness measure.

The purpose of this paper is to investigate roughness measure in an axiomatic way. We first introduce an axiomatic definition of roughness measure by taking into account the common properties of the roughness measures from [35] and [17]. After giving an axiomatic definition of partition measure, we then provide a unified construction of roughness measure, called strong Pawlak roughness measure, by combining partition measure into Pawlak's roughness measure. Some properties of the strong Pawlak roughness measure are examined in detail. Finally, we show that the existing roughness measures in [35] and [17] are two special instances of our strong Pawlak roughness measure and present three new strong Pawlak roughness measures as well. The advantage of our axiomatic approach is that some properties of a roughness measure follow immediately as soon as the measure satisfies the relevant axiomatic definition.

The remainder of the paper is structured as follows. In Section 2, we briefly review some basics of Pawlak's rough set theory and the roughness measures in the literature. The axiomatic definitions of roughness measure and partition measure are given in Section 3. The strong Pawlak roughness measure and its properties are also provided in this section. Section 4 is devoted to the case study of strong Pawlak roughness measures. We conclude the paper in Section 5 with a brief discussion on the future research.

## 2. Preliminaries

This section consists of four subsections. We recall the definition of Pawlak's rough sets in Section 2.1. Sections 2.2, 2.3, and 2.4 are devoted to roughness measures proposed by Pawlak [23], Xu et al. [35], and Liang et al. [17], respectively. Some necessary properties of these measures are collected for later use.

*2.1. Rough sets*

We start by recalling some basic notions in Pawlak's rough set theory [22, 23].

Let $U$ be a finite and nonempty universal set, and let $R \subseteq U \times U$ be an equivalence relation on $U$. Denote by $U/R$ the set of all equivalence classes induced by $R$. Such equivalence classes are also called *elementary sets*; every union (not necessarily nonempty) of elementary sets is called a *definable set*. For any $A \subseteq U$, one can characterize $A$ by a pair of lower and upper approximations. The *lower approximation* $R_*(A)$ of $A$ is defined as the greatest definable set contained in $A$, while the *upper approximation* $R^*(A)$ of $A$ is defined as the least definable set containing $A$. Formally,

$$R_*(A) = \cup \{C \in U/R \mid C \subseteq A\} \text{ and } R^*(A) = \cup \{C \in U/R \mid C \cap A \neq \emptyset\}.$$

It follows immediately from definition that $R_*(A) \subseteq A \subseteq R^*(A)$ for any $A \subseteq U$. In particular, when $R_*(A) = A = R^*(A)$, the set $A$ is also called *R-exact*. Clearly, every definable set is $R$-exact, and vice versa. The ordered pair $\langle U, R \rangle$ is said to be an *approximation space*. A *rough set* in $\langle U, R \rangle$ is the family of subsets of $U$ with the same lower and upper approximations.

Recall that a *partition* of $U$ is a collection of nonempty subsets of $U$ such that every element $x$ in $U$ is in exactly one of these subsets. We write $\Pi(U)$ for the set of all partitions of $U$ and $\mathscr{P}(U)$ for the power set of $U$. It is well-known that the notions partition and equivalence relation are essentially equivalent, that is, for any equivalence relation $R$ on $U$, the set $U/R$ is a partition of $U$, and conversely, from any partition $\pi$ of $U$, one can define an equivalence relation $R_\pi$ on $U$ such that $U/R_\pi = \pi$ in the obvious way. Thus, we sometimes say that the ordered pair $\langle U, \pi \rangle$ is an approximation space and write $\pi_*(A)$ and $\pi^*(A)$ for $R_{\pi*}(A)$ and $R_\pi^*(A)$, respectively. More generally, we will use equivalence relation and partition indiscriminately.

Whatever be a nonempty universe $U$, it is always possible to introduce at least two canonical partitions: One is the trivial partition, denoted by $\check{\pi}$, consisting of a unique equivalence class, and the other is the discrete partition, denoted by $\hat{\pi}$, consisting of all singletons from $U$. Formally,

$$\check{\pi} = \{U\} \text{ and } \hat{\pi} = \{\{x\} \mid x \in U\}.$$

We now define a partial order "≤" on $\Pi(U)$: For any $\pi, \sigma \in \Pi(U)$,

$$\pi \leq \sigma \iff \text{for any } C \in \pi, \text{ there exists } D \in \sigma \text{ such that } C \subseteq D.$$

For instance, $\hat{\pi} \leq \pi \leq \check{\pi}$ for any $\pi \in \Pi(U)$. We say that $\pi$ is *finer* than $\sigma$ and that $\sigma$ is *coarser* than $\pi$ if $\pi \leq \sigma$. When $\pi \prec \sigma$, that is, $\pi \leq \sigma$ and $\pi \neq \sigma$, we say that $\pi$ is *strictly finer* than $\sigma$ and that $\sigma$ is *strictly coarser* than $\pi$. Informally, this means that $\pi$ is a further fragmentation of $\sigma$.

*2.2. Roughness measure by Pawlak*

Let $\langle U, \pi \rangle$ be an approximation space. To characterize the uncertainty of rough sets, Pawlak proposed two numerical measures: roughness and accuracy (see, for example, [23]). The accuracy reflects the degree of completeness of knowledge about a given subset $A$ of $U$; it is defined by the ratio of the cardinalities of the lower and upper approximations. More formally, the *accuracy* of $A$ (with respect to $\pi$) is defined by

$$\alpha_P(\pi, A) = \frac{|\pi_*(A)|}{|\pi^*(A)|},$$

where $A \neq \emptyset$ and "$|S|$" denotes the cardinality of a set $S$. For convenience, we set $\alpha_P(\pi, \emptyset) = 1$, that is, $|\emptyset|/|\emptyset| = 1$.

The roughness is opposed to the accuracy; it represents the degree of incompleteness of knowledge about a given subset $A$. The *roughness* $\beta_P(\pi, A)$ of $A$ (with respect to $\pi$) is calculated by subtracting the accuracy $\alpha_P(\pi, A)$ from 1, that is,

$$\beta_P(\pi, A) = 1 - \alpha_P(\pi, A).$$

Clearly, $0 \leq \beta_P(\pi, A) \leq 1$. It is easy to see that $\beta_P(\pi, A) = 1$ if and only if $\pi_*(A) = \emptyset$ and $\pi^*(A) \neq \emptyset$, and $\beta_P(\pi, A) = 0$ if and only if $A$ is $\pi$-exact.

To state the next property, we need one more notation. By $\beta_P(\pi, \cdot) \lneq \beta_P(\sigma, \cdot)$ we mean that $\beta_P(\pi, A) \leq \beta_P(\sigma, A)$ for every $A \subseteq U$ and $\beta_P(\pi, B) \neq \beta_P(\sigma, B)$ for some $B \subseteq U$. Similar usages of "$\lneq$" will appear in the subsequent sections. The following property follows from the fact that for any $A \subseteq U$, $\sigma_*(A) \subseteq \pi_*(A) \subseteq A \subseteq \pi^*(A) \subseteq \sigma^*(A)$ if $\pi \leq \sigma$.

**Property 1.** *For any $\pi, \sigma \in \Pi(U)$, if $\pi \prec \sigma$, then $\beta_P(\pi, \cdot) \lneq \beta_P(\sigma, \cdot)$.*

The roughness and accuracy measures are of simple expressions and are suitable for evaluating the uncertainty arising from the boundary region. Nevertheless, neither the roughness nor the accuracy reflects the granularity of the underlying partition. This limitation has been pointed out by several researchers [2, 17, 35]. For the convenience of the reader, we record a simple example.

**Example 1.** *Let $U = \{a_1, a_2, a_3, a_4, a_5\}$, $\pi = \{\{a_1\}, \{a_2, a_3\}, \{a_4, a_5\}\}$, and $\sigma = \{\{a_1, a_2, a_3\}, \{a_4, a_5\}\}$. Then it is easy to see that $\pi$ and $\sigma$ are partitions of $U$, and moreover, $\pi \prec \sigma$. Assume that $A = \{a_1, a_2, a_3, a_4\}$. We thus have by definition that*

$$\pi_*(A) = \sigma_*(A) = \{a_1, a_2, a_3\} \text{ and } \pi^*(A) = \sigma^*(A) = \{a_1, a_2, a_3, a_4, a_5\} = U.$$

*Therefore, $\alpha_P(\pi, A) = \alpha_P(\rho, A) = 0.6$ and $\beta_P(\pi, A) = \beta_P(\rho, A) = 0.4$. It means that partitions with different granulations may give rise to the same accuracy (and thus roughness) for some subsets. This property is, of course, not desirable.*

*2.3. Roughness measure by Xu et al.*

To improve the accuracy and roughness measures suggested by Pawlak, Xu et al. [35] combined the granularity of the underlying partition into Pawlak's accuracy by exploiting the so-called equivalence relation graphs. To introduce their definition, we need several notions.

Let $\langle U, \pi \rangle$ be an approximation space. The *equivalence relation graph* with respect to $\pi$ is defined as $G(\pi) = (N(\pi), E(\pi))$, where $N(\pi) = U$ and $E(\pi) = \{(x, y) \in U \times U \mid (x, y) \in R_\pi\}$. For any vertex $v$ on $G(\pi) = (N(\pi), E(\pi))$, let the subgraph corresponding to $v$ be $G(\pi)^{(v)} = (N(\pi)^{(v)}, E(\pi)^{(v)})$, where $N(\pi)^{(v)} = N(\pi)$ and $E(\pi)^{(v)} = \{(v, y) \mid (v, y) \in E(\pi)\}$. For any given $G(\pi)$, let $p_{L(G(\pi), v)}^{G(\pi)}$ be the proportion of the row vector $L(G(\pi), v)$ out of $|N(\pi)|$ row vectors in the incidence matrix of $G(\pi)$, where $L(G(\pi), v)$ is the row vector corresponding to $v$ in the incidence matrix of $G(\pi)$. More formally, $p_{L(G(\pi),v)}^{G(\pi)} = n/|N(\pi)|$, where $n$ is the number of row vectors in the incidence matrix of $G(\pi)$ that are equal to $L(G(\pi), v)$. For an equivalence relation graph $G(\pi)$, the *minimum description length* of $G(\pi)$ is defined by[1]

$$I(G(\pi)) = -\sum_{v \in N(\pi)} \log p_{L(G(\pi),v)}^{G(\pi)}.$$

Finally, we may state the key definition and results in [35] as follows.

Let $\langle U, \pi \rangle$ be an approximation space. The *roughness* of $A$ (with respect to $\pi$) due to Xu et al. is defined by

$$\beta_X(\pi, A) = \beta_P(\pi, A) \cdot \frac{con(G(\pi))}{|U|(|U|-1)\log|U|},$$

where $\beta_P(\pi, A)$ is the Pawlak's roughness of $A$ and

$$con(G(\pi)) = \sum_{v \in N(\pi)} I(G(\pi)^{(v)}) - I(G(\pi)).$$

Note that $I(G(\pi)^{(v)})$ is the minimum description length of the subgraph $G(\pi)^{(v)}$ corresponding to $v$.

It follows from [35] that the new roughness measure $\beta_X(\pi, A)$ enjoys some useful properties, which are listed as follows.

**Property 2.** For any $\pi \in \Pi(U)$ and $A \subseteq U$, $0 \le \beta_X(\pi, A) \le 1$.

**Property 3.** $\beta_X(\hat{\pi}, A) = 0$ for any $A \subseteq U$; $\beta_X(\check{\pi}, U) = 0$, and $\beta_X(\check{\pi}, A) = 1$ for any $A \subsetneq U$.

**Property 4.** For any $\pi, \sigma \in \Pi(U)$, if $\pi \le \sigma$, then $\beta_X(\pi, A) \le \beta_X(\sigma, A)$ for any $A \subseteq U$.

**Property 5.** Let $\pi \in \Pi(U)$ and $A, B \subseteq U$.

1) If $\pi_*(A) = \pi_*(B)$, then $\beta_X(\pi, A \cap B) \le \min\{\beta_X(\pi, A), \beta_X(\pi, B)\}$.
2) If $\pi^*(A) = \pi^*(B)$, then $\beta_X(\pi, A \cup B) \le \min\{\beta_X(\pi, A), \beta_X(\pi, B)\}$.

It should be pointed out that $\beta_X$ has not the same drawback as shown for $\beta_P$ in Example 1.

2.4. *Roughness measure by Liang et al.*

Based on the notion of knowledge granulation due to Miao and Fan [21], Liang et al. proposed a roughness measure [17] which is simpler than that in [35]. Let $\langle U, \pi \rangle$ be an approximation space and suppose that $\pi = \{C_1, C_2, \ldots, C_m\}$. The *roughness* of $A$ (with respect to $\pi$) due to Liang et al. is defined by

$$\beta_L(\pi, A) = \beta_P(\pi, A) \cdot \frac{\sum_{i=1}^m |C_i|^2}{|U|^2},$$

where $\beta_P(\pi, A) = 1 - |\pi_*(A)|/|\pi^*(A)|$ is the Pawlak's roughness of $A$ and the second term $\frac{\sum_{i=1}^m |C_i|^2}{|U|^2}$ represents the knowledge granulation of $\pi$ introduced in [21].

The following properties of the roughness measure $\beta_L(\pi, A)$ were given in [17].

**Property 6.** For any $\pi \in \Pi(U)$ and $A \subseteq U$, $0 \le \beta_L(\pi, A) \le 1$.

**Property 7.** If $\beta_L(\pi, A) = 0$ for all $A \subseteq U$, then $\pi = \hat{\pi}$; if $\beta_L(\pi, A) = 1$ for some $A \ne U$, then $\pi = \check{\pi}$.

**Property 8.** For any $\pi, \sigma \in \Pi(U)$, if $\pi \le \sigma$, then $\beta_L(\pi, A) \le \beta_L(\sigma, A)$ for any $A \subseteq U$.

Like $\beta_X$, the roughness measure $\beta_L$ has not the same drawback as shown for $\beta_P$ in Example 1.

---

[1] All logarithms are to base 2 unless otherwise specified.

## 3. An axiomatic definition of roughness measure

This section is composed of three subsections. Motivated by the roughness measures in [17, 35], we present an axiomatic definition of roughness measure and discuss its basic properties in Section 3.1. In order to construct more roughness measures, we introduce an axiomatic definition of partition measure in Section 3.2. By incorporating partition measure into Pawlak's roughness measure, we introduce a strong Pawlak roughness measure and explore its properties in Section 3.3.

*3.1. Roughness measure*

For later need, let us introduce the following notion which relates two approximation spaces.

**Definition 1.** *Let $\langle U, \pi \rangle$ and $\langle V, \sigma \rangle$ be two approximation spaces, and suppose that $f : U \longrightarrow V$ is a mapping.*

1) *The mapping $f$ is called a* homomorphism *from $\langle U, \pi \rangle$ to $\langle V, \sigma \rangle$ if for any $C \in \pi$, there exists $D \in \sigma$ such that $f(C) \subseteq D$, where $f(C) = \{f(u) \mid u \in C\}$.*
2) *A homomorphism $f$ is called a* monomorphism *if $f$ is an injective mapping.*
3) *A monomorphism $f$ is called* strictly monomorphic *if there exist $C \in \pi$ and $D \in \sigma$ such that $f(C) \subsetneq D$, namely, $f(C) \subseteq D$ and $f(C) \neq D$.*
4) *The mapping $f$ is called an* isomorphism *if the mapping $f : U \longrightarrow V$ is bijective, and moreover, both $f$ and its inverse mapping $f^{-1}$ are homomorphisms.*

Let us now present the axiomatic definition of roughness measure.

**Definition 2.** *Let $U$ be a finite and nonempty universal set and $\beta$ a mapping from $\Pi(U) \times \mathscr{P}(U)$ to the closed unit interval $[0, 1]$. We say that $\beta$ is a* roughness measure *on $U$ if the following conditions are satisfied:*

1) *$\beta(\pi, A) = 0$ if and only if $A$ is $\pi$-exact.*
2) *For any $\pi, \sigma \in \Pi(U)$, if $\pi \prec \sigma$, then $\beta(\pi, \cdot) \precsim \beta(\sigma, \cdot)$.*
3) *For any $\pi, \sigma \in \Pi(U)$, if there is an isomorphism $f$ from $\langle U, \pi \rangle$ to $\langle U, \sigma \rangle$, then $\beta(\pi, A) = \beta(\sigma, f(A))$ for any $A \subseteq U$.*

*If $\beta$ is a roughness measure on $U$, then the* roughness *of $A$ (with respect to $\pi$) is defined by the value $\beta(\pi, A)$.*

Let us give a brief, informal account of the above conditions. Condition 1) just says that a set is exact if and only if it has roughness 0. Condition 2) requires that roughness measure strictly maintains the partial order on $\Pi(U)$. Notice that any isomorphism $f$ from $\langle U, \pi \rangle$ to $\langle U, \sigma \rangle$ is actually a renaming of elements of $U$ that keeps elementary sets. For example, there is an isomorphism between $\langle \{1, 2, 3, 4\}, \{\{1, 2\}, \{3, 4\}\} \rangle$ and $\langle \{1, 2, 3, 4\}, \{\{1, 3\}, \{2, 4\}\} \rangle$. Therefore, the condition 3) requires that roughness measure is only dependent on the structure (i.e., blocks) of partitions; this seems quite reasonable.

It follows from the definition of Pawlak's roughness measure $\beta_P$ and Property 1 that $\beta_P$ is a roughness measure in the sense of Definition 2. To illustrate the definition, let us examine a trivial example.

**Example 2.** *Consider $\beta : \Pi(U) \times \mathscr{P}(U) \longrightarrow [0, 1]$ defined as follows:*

$$\beta(\pi, A) = \begin{cases} 0, & \text{if } A \text{ is } \pi\text{-exact} \\ 1, & \text{otherwise.} \end{cases}$$

*Clearly, the condition 1) in Definition 2 is satisfied. For the condition 2), let $\pi, \sigma \in \Pi(U)$ with $\pi \prec \sigma$. For any $A \subseteq U$, it follows from definition that $A$ is $\pi$-exact whenever it is $\sigma$-exact, but the converse does not hold, that is, a $\pi$-exact set may not be $\sigma$-exact. It means that if $\beta(\sigma, A) = 0$, then $\beta(\pi, A) = 0$, and there exists $A' \subseteq U$ such that $\beta(\pi, A') = 0$ while $\beta(\sigma, A') = 1$. This forces that $\beta(\pi, \cdot) \precsim \beta(\sigma, \cdot)$, as desired. For the condition 3), note that the isomorphism $f$ establishes a one-to-one correspondence between the set of $\pi$-exact sets and that of $\sigma$-exact ones. Therefore, the condition 3) holds, and $\beta$ is indeed a roughness measure on $U$.*

By definition, we have two properties of roughness measure. The first one is a characterization of the minimum of a roughness measure.

**Proposition 1.** *Let $\beta$ be a roughness measure on $U$. Then $\beta(\pi, A) = 0$ holds for all $A \subseteq U$ if and only if $\pi = \hat{\pi}$.*

PROOF. Note that if $\pi = \hat{\pi}$, then every subset $A$ of $U$ is $\pi$-exact. Hence, the sufficiency follows immediately from Definition 2. Conversely, if $\beta(\pi, A) = 0$ holds for all $A \subseteq U$, then by definition every subset $A$ of $U$ is $\pi$-exact. This means that every nonempty subset of $U$ is a definable set. As a result, we see that $\pi = \hat{\pi}$, which proves the necessity.

The next property is a relaxation of the condition 2) in Definition 2.

**Proposition 2.** *Let $\beta$ be a roughness measure on $U$ and $\pi, \sigma \in \Pi(U)$. If $\pi \leq \sigma$, then $\beta(\pi, A) \leq \beta(\sigma, A)$ for any $A \subseteq U$. In particular, $\beta(\pi, A) \leq \beta(\check{\pi}, A)$ for any $\pi \in \Pi(U)$ and $A \subseteq U$.*

PROOF. It follows directly from the condition 2) in Definition 2.

The following property is equivalent to the condition 2) in Definition 2, under the condition 3) in this definition.

**Proposition 3.** *Suppose that $\beta$ is a roughness measure on $U$ and $f$ is a strict monomorphism from $\langle U, \pi \rangle$ to $\langle U, \sigma \rangle$. Then $\beta(\pi, A) \leq \beta(\sigma, f(A))$ for any $A \subseteq U$, and moreover, there exists $A' \subseteq U$ such that $\beta(\pi, A') < \beta(\sigma, f(A'))$.*

PROOF. Since $f$ is a monomorphism from $\langle U, \pi \rangle$ to $\langle U, \sigma \rangle$, it gives an isomorphism between $\langle U, \pi \rangle$ and $\langle U, f(\pi) \rangle$, where $f(\pi) = \{f(A) \mid A \in \pi\}$. Note that $U$ is finite and $f$ is injective, so $f$ is bijective and thus $f(\pi)$ is indeed a partition of $U$. By definition, we see that $\beta(\pi, A) \leq \beta(f(\pi), f(A))$ for any $A \subseteq U$. On the other hand, we have that $f(\pi) \prec \sigma$ since $f$ is strictly monomorphic. This means by Proposition 2 that $\beta(f(\pi), f(A)) \leq \beta(\sigma, f(A))$ for any $A \subseteq U$. As a result, $\beta(\pi, A) \leq \beta(\sigma, f(A))$ for any $A \subseteq U$. The remainder of this proposition follows easily from the strictness of the monomorphism $f$.

The condition 2) in Definition 2 just says that $\beta$ is strictly monotonic. Depending on applications, strict monotonicity may not be so required. For example, it is usually interesting to look at dependencies between partitions generated by decision attributes and condition attributes in a decision system. In such cases, one may be mostly interested in weak monotonicity but not in strict monotonicity. Actually, the cases when the measure does not change while changing the partition into a more or less detailed one are of special importance for feature selection, feature subset selection, feature extraction, and feature reduction in knowledge discovery (see, for example, [7, 11, 31]). In view of this, let us introduce a weak version of Definition 2 as follows.

**Definition 3.** *Let $U$ be a finite and nonempty universal set and $\beta$ a mapping from $\Pi(U) \times \mathcal{P}(U)$ to the closed unit interval $[0, 1]$. We say that $\beta$ is a weak roughness measure on $U$ if the following conditions are satisfied:*

1) $\beta(\pi, A) = 0$ if and only if $A$ is $\pi$-exact.
2) *For any $\pi, \sigma \in \Pi(U)$, if $\pi \leq \sigma$, then $\beta(\pi, \cdot) \leq \beta(\sigma, \cdot)$.*
3) *For any $\pi, \sigma \in \Pi(U)$, if there is an isomorphism $f$ from $\langle U, \pi \rangle$ to $\langle U, \sigma \rangle$, then $\beta(\pi, A) = \beta(\sigma, f(A))$ for any $A \subseteq U$.*
4) $\beta(\pi, \cdot) = 0$ *if and only if* $\pi = \hat{\pi}$.

*If $\beta$ is a weak roughness measure on $U$, then the* weak roughness *of $A$ (with respect to $\pi$) is defined by the value $\beta(\pi, A)$.*

Clearly, Conditions 1) and 3) are the same as in Definition 2, and Condition 2) means that $\beta$ is weakly monotonic. Note that in order to avoid $\beta$ being constant, we add Condition 4) in the above definition. It is easy to check that any roughness measure in the sense of Definition 2 is a weak roughness measure, but the converse does not hold in general.

*3.2. Partition measure*

In order to measure partitions, we provide an axiomatic definition of partition measure as follows.

**Definition 4.** *Let $U$ be a finite and nonempty universal set and $h$ a mapping from $\Pi(U)$ to $[0, +\infty)$, the set of nonnegative real numbers. We say that $h$ is a* partition measure *on $U$ if the following conditions are satisfied:*

1) *For any $\pi, \sigma \in \Pi(U)$, if $\pi \prec \sigma$, then $h(\pi) < h(\sigma)$.*

2) *For any $\pi, \sigma \in \Pi(U)$, if there is an isomorphism $f$ from $\langle U, \pi \rangle$ to $\langle U, \sigma \rangle$, then $h(\pi) = h(\sigma)$.*

Intuitively, we require that partition measures on $U$ are only dependent upon the structure of partitions, not the names of elements in $U$. Roughly speaking, the greater the value of $h$, the coarser the corresponding partition. Let us see an example.

**Example 3.** *Let $U = \{1, 2, 3, 4\}$. Then $U$ has 15 partitions because the total number of partitions of an n-element set is the Bell number $B_n$, recursively defined by $B_{n+1} = \sum_{k=0}^{n} \binom{n}{k} B_k$ and $B_0 = 1$ (see, for example, [5]). For simplicity, we write $1/2/34$ for the partition $\{\{1\}, \{2\}, \{3, 4\}\}$, alike to other partitions. With this notation, we have that*

$$\Pi(U) = \{1234, 1/234, 2/134, 3/124, 4/123, 14/23, 13/24, 12/34,$$
$$1/2/34, 1/3/24, 1/4/23, 3/4/12, 2/4/13, 2/3/14, 1/2/3/4\}.$$

*By definition, any mapping $h : \Pi(U) \longrightarrow [0, +\infty)$ that satisfies the following conditions is a partition measure on $U$:*

$$h(1/2/3/4) = r_1,$$
$$h(1/2/34) = h(1/3/24) = h(1/4/23) = h(3/4/12)$$
$$= h(2/4/13) = h(2/3/14) = r_2,$$
$$h(14/23) = h(13/24) = h(12/34) = r_3,$$
$$h(1/234) = h(2/134) = h(3/124) = h(4/123) = r_4,$$
$$h(1234) = r_5,$$

*where $r_i \in [0, +\infty)$, $i = 1, 2, \ldots, 5$, with $r_1 < r_2 < r_3 < r_5$ and $r_1 < r_2 < r_4 < r_5$.*

We remark that our axiomatic definition of partition measure is essentially based on the cardinalities of all equivalence classes in a partition. Recently, Yao and Zhao [41] have directly established a partition measure on the cardinality of a partition, and moreover, they constructed an interesting measure of the granularity of a partition which has several existing measures as instances. We see by Theorem 3 in [41] that this new measure satisfies Definition 4 as well. It should be stressed that constructing and evaluating partitions are the most basic issues in rough set theory, since the indiscernibility is the mathematical basis of rough set theory [25] and there are strong relationships between indiscernibility measures and partition measures [1, 21, 32, 39, 40, 45]. In addition to rough sets, the granularity of a partition is a very important concept in many other fields such as information theory, data mining, machine learning, and pattern recognition. In the literature, there are a large number of approaches to measuring partitions (see, for example, [2, 9, 10, 12, 13, 14, 15, 16, 17, 18, 24, 25, 29, 33, 34, 36, 38, 39, 42, 47]). In some sense, the roughness measure in Definition 2 as well as the weak roughness measure in Definition 3 can be viewed as a partition measure because for some given subsets, say $A$, of $U$, the function $\beta(\cdot, A)$ can reflect the granularity of a partition.

The following facts follow directly from Definition 4.

**Corollary 1.** *Suppose that $h$ is a partition measure on $U$.*

1) *For any $\pi, \sigma \in \Pi(U)$, if $\pi \preceq \sigma$, then $h(\pi) \leq h(\sigma)$.*
2) *For any $\pi \in \Pi(U)$, $h(\hat{\pi}) \leq h(\pi) \leq h(\check{\pi})$. In particular, $h(\pi) > 0$ whenever $\pi \neq \hat{\pi}$.*

As expected, partition measures have the following property.

**Proposition 4.** *Let $h$ be a partition measure on $U$ and $\pi, \sigma \in \Pi(U)$. If there is a strict monomorphism $f$ from $\langle U, \pi \rangle$ to $\langle U, \sigma \rangle$, then $h(\pi) < h(\sigma)$.*

PROOF. Assume that $f$ is a strict monomorphism from $\langle U, \pi \rangle$ to $\langle U, \sigma \rangle$. Then it is clear that $f$ gives rise to an isomorphism from $\langle U, \pi \rangle$ to $\langle U, f(\pi) \rangle$, where $f(\pi) = \{f(A) \mid A \in \pi\}$. We thus have that $h(\pi) = h(f(\pi))$ by Definition 4. Furthermore, we find that $f(\pi) \prec \sigma$ since $f$ is strictly monomorphic. By definition, we obtain that $h(f(\pi)) < h(\sigma)$. Consequently, $h(\pi) < h(\sigma)$, finishing the proof.

As a corollary of Proposition 4, we get an equivalent definition of partition measure.

**Corollary 2.** *A mapping $h : \Pi(U) \longrightarrow [0, +\infty)$ is a partition measure on $U$ if and only if the following conditions hold:*

1) *For any $\pi, \sigma \in \Pi(U)$, if there is a strict monomorphism $f$ from $\langle U, \pi \rangle$ to $\langle U, \sigma \rangle$, then $h(\pi) < h(\sigma)$.*
2) *For any $\pi, \sigma \in \Pi(U)$, if there is an isomorphism $f$ from $\langle U, \pi \rangle$ to $\langle U, \sigma \rangle$, then $h(\pi) = h(\sigma)$.*

*3.3. Strong Pawlak roughness measure*

Given a partition measure, we can construct a roughness measure in the sense of Definition 2 as follows.

**Theorem 1.** *Let $U$ be a finite and nonempty universal set. Suppose that $h$ is a partition measure on $U$. Then the function $\beta_h : \Pi(U) \times \mathscr{P}(U) \longrightarrow [0, 1]$ defined by*

$$\beta_h(\pi, A) = \beta_P(\pi, A) \cdot \frac{h(\pi)}{h(\check{\pi})} = \left(1 - \frac{|\pi_*(A)|}{|\pi^*(A)|}\right) \cdot \frac{h(\pi)}{h(\check{\pi})}$$

*is a roughness measure on $U$.*

PROOF. We need to check all the three conditions in Definition 2.

For the first condition, observe that $\beta_h(\pi, A) = 0$ if and only if either $\frac{|\pi_*(A)|}{|\pi^*(A)|} = 1$ or $h(\pi) = 0$. Clearly, $\frac{|\pi_*(A)|}{|\pi^*(A)|} = 1$ is equivalent to that $A$ is $\pi$-exact. Note that $h(\pi) = 0$ implies $\pi = \hat{\pi}$ by Corollary 1. In this case, every subset $A$ of $U$ is $\pi$-exact. Hence, $\beta_h(\pi, A) = 0$ if and only if $A$ is $\pi$-exact, as desired.

For the second condition, assume that $\pi \prec \sigma$. Then $h(\pi) < h(\sigma)$ by definition. On the other hand, we have by Property 1 that $\beta_P(\pi, \cdot) \lneq \beta_P(\sigma, \cdot)$. We thus obtain that $\beta_h(\pi, \cdot) \lneq \beta_h(\sigma, \cdot)$, as desired.

For the third condition, suppose that $f$ is an isomorphism from $\langle U, \pi \rangle$ to $\langle U, \sigma \rangle$, where $\pi, \sigma \in \Pi(U)$. Then by definition we see that $|\pi_*(A)| = |\sigma_*(f(A))|$ and $|\pi^*(A)| = |\sigma^*(f(A))|$ for any $A \subseteq U$. Therefore, $\beta_P(\pi, A) = \beta_P(\sigma, f(A))$. Since $f$ is an isomorphism, we also have that $h(\pi) = h(\sigma)$ by definition. As a result, we get that $\beta_h(\pi, A) = \beta_h(\sigma, f(A))$ for any $A \subseteq U$. This completes the proof of the theorem.

For convenience, the roughness measure $\beta_h$ associated to a partition measure $h$ is called a *strong Pawlak roughness measure*. Note that neither Pawlak's roughness measure $\beta_P$ nor the roughness measure defined in Example 2 is a strong Pawlak roughness measure. We end this subsection with a discussion on the properties of strong Pawlak roughness measures. The first five properties follow immediately from Theorem 1, in which the assertion 3) justifies the modifier "strong".

**Corollary 3.**

1) $\beta_h(\pi, A) = 0$ *if and only if $A$ is $\pi$-exact.*
2) *For any $\pi, \sigma \in \Pi(U)$, if $\pi \prec \sigma$, then $\beta_h(\pi, \cdot) \lneq \beta_h(\sigma, \cdot)$. Consequently, if $\pi \leq \sigma$, then $\beta_h(\pi, A) \leq \beta_h(\sigma, A)$ for any $A \subseteq U$.*
3) *For any $\pi, \sigma \in \Pi(U)$, if $\pi \prec \sigma$ and $A \subseteq U$ is not $\sigma$-exact, then $\beta_h(\pi, A) < \beta_h(\sigma, A)$.*
4) *For any $\pi, \sigma \in \Pi(U)$, if there is an isomorphism $f$ from $\langle U, \pi \rangle$ to $\langle U, \sigma \rangle$, then $\beta_h(\pi, A) = \beta_h(\sigma, f(A))$ for any $A \subseteq U$.*
5) *Assume that there is a strict monomorphism $f$ from $\langle U, \pi \rangle$ to $\langle U, \sigma \rangle$. Then $\beta_h(\pi, A) \leq \beta_h(\sigma, f(A))$ for any $A \subseteq U$, and moreover, there exists $A' \subseteq U$ such that $\beta_h(\pi, A') < \beta_h(\sigma, f(A'))$.*

Like the previous roughness measures $\beta_P$, $\beta_X$, and $\beta_L$, the strong Pawlak roughness measure $\beta_h$ is bounded as well.

**Proposition 5.** *For any $\pi \in \Pi(U)$ and $A \subseteq U$, $0 \leq \beta_h(\pi, A) \leq 1$.*

PROOF. It follows directly from $0 \leq \beta_P(\pi, A) \leq 1$ and Corollary 1.

In terms of the maximum and minimum of $\beta_h$, we have further characterizations.

**Proposition 6.** $\beta_h(\pi, A) = 0$ *holds for all $A \subseteq U$ if and only if $\pi = \hat{\pi}$.*

PROOF. By the assertion 1) in Corollary 3, $\beta_h(\pi, A) = 0$ holds for all $A \subseteq U$ if and only if every subset $A$ of $U$ is $\pi$-exact, which is equivalent to that $\pi = \hat{\pi}$. This proves the proposition.

**Proposition 7.** *If there exists $A \subseteq U$ such that $\beta_h(\pi, A) = 1$, then $\pi = \check{\pi}$. Moreover,*

$$\beta_h(\check{\pi}, A) = \begin{cases} 0, & \text{if } A = U \\ 1, & \text{otherwise.} \end{cases}$$

Proof. If there exists $A \subseteq U$ such that $\beta_h(\pi, A) = 1$, then it forces that $\beta_P(\pi, A) = 1$ and $h(\pi) = h(\check{\pi})$. The latter yields that $\pi = \check{\pi}$ by Corollary 1. The remainder of this proposition follows readily from the definition of $\beta_h$.

Like the roughness measure $\beta_X$, the strong Pawlak roughness measure $\beta_h$ has the following properties.

**Proposition 8.** *Let $\pi \in \Pi(U)$ and $A, B \subseteq U$.*
  1) *If $\pi_*(A) = \pi_*(B)$, then $\beta_h(\pi, A \cap B) \leq \min\{\beta_h(\pi, A), \beta_h(\pi, B)\}$.*
  2) *If $\pi^*(A) = \pi^*(B)$, then $\beta_h(\pi, A \cup B) \leq \min\{\beta_h(\pi, A), \beta_h(\pi, B)\}$.*

Proof. 1) Clearly, $\pi^*(A \cap B) \subseteq \pi^*(A)$. If $\pi_*(A) = \pi_*(B)$, then we can verify that $\pi_*(A \cap B) = \pi_*(A)$. Hence,

$$\frac{|\pi_*(A \cap B)|}{|\pi^*(A \cap B)|} \geq \frac{|\pi_*(A)|}{|\pi^*(A)|},$$

which means that $\beta_P(\pi, A \cap B) \leq \beta_P(\pi, A)$. In the same way, we obtain that $\beta_P(\pi, A \cap B) \leq \beta_P(\pi, B)$. It yields that $\beta_h(\pi, A \cap B) \leq \min\{\beta_h(\pi, A), \beta_h(\pi, B)\}$, as desired.

2) The proof is similar to that of 1). It is easy to see that $\pi_*(A) \subseteq \pi_*(A \cup B)$. If $\pi^*(A) = \pi^*(B)$, then we can check that $\pi^*(A \cup B) = \pi^*(A)$. As a result, we have that

$$\frac{|\pi_*(A \cup B)|}{|\pi^*(A \cup B)|} \geq \frac{|\pi_*(A)|}{|\pi^*(A)|}.$$

Therefore, $\beta_P(\pi, A \cup B) \leq \beta_P(\pi, A)$. Analogously, we can obtain that $\beta_P(\pi, A \cup B) \leq \beta_P(\pi, B)$. It gives that $\beta_h(\pi, A \cup B) \leq \min\{\beta_h(\pi, A), \beta_h(\pi, B)\}$, finishing the proof of the proposition.

## 4. Case study of strong Pawlak roughness measures

As we have seen in the previous section, if one can show that $h$ is a partition measure on $U$, then $\beta_h$ is a strong Pawlak roughness measure and thus has the properties stated in Corollary 3 and Propositions 5, 6, 7, and 8. In this section, we first look at the roughness measures $\beta_X$ and $\beta_L$ due to Xu et al. [35] and Liang et al. [17], respectively, in the framework of strong Pawlak roughness measures, and then provide three new strong Pawlak roughness measures.

**Proposition 9.** *The roughness measure $\beta_X$ due to Xu et al. is a strong Pawlak roughness measure.*

Proof. By definition,
$$\beta_X(\pi, A) = \beta_P(\pi, A) \cdot \frac{con(G(\pi))}{|U|(|U|-1)\log|U|},$$

so it is sufficient to show that $h$ defined by

$$h(\pi) = \frac{con(G(\pi))}{|U|(|U|-1)\log|U|}$$

is a partition measure on $U$ and $h(\check{\pi}) = 1$. It follows from Theorem 7 in [35] that $h$ satisfies the condition 1) in Definition 4, that is, $\pi \prec \sigma$ implies $h(\pi) < h(\sigma)$. In addition, if there is an isomorphism from $\langle U, \pi \rangle$ to $\langle U, \sigma \rangle$, then we see that $con(G(\pi)) = con(G(\sigma))$ by the definition of equivalence relation graph. Therefore, $h(\pi) = h(\sigma)$. It follows from definition that $con(G(\check{\pi})) = |U|(|U|-1)\log|U|$, namely, $h(\check{\pi}) = 1$. Hence, $h$ is a partition measure, as desired. In fact, we may also define $h'(\pi) = con(G(\pi))$ and then verify that $h'$ is a partition measure and $h'(\check{\pi}) = |U|(|U|-1)\log|U|$.

**Proposition 10.** *The roughness measure $\beta_L$ due to Liang et al. is a strong Pawlak roughness measure.*

Proof. By definition,
$$\beta_L(\pi, A) = \beta_P(\pi, A) \cdot \frac{\sum_{i=1}^{m}|C_i|^2}{|U|^2},$$

where $\{C_1, C_2, \ldots, C_m\} = \pi$. In order to show that $\beta_L$ is a strong Pawlak roughness measure, it suffices to prove that $h$ defined by $h(\pi) = \sum_{i=1}^{m}|C_i|^2$ is a partition measure on $U$ and $h(\check{\pi}) = |U|^2$. The former follows readily from the definition of $h$ and the latter is obvious. Therefore, $\beta_L$ is a strong Pawlak roughness measure.

**Remark 1.** Propositions 9 and 10 tell us that the roughness measures $\beta_X$ and $\beta_L$ have the properties stated in Corollary 3 and Propositions 5, 6, 7, and 8; some of these properties are missing in [35] or [17].

To introduce another strong Pawlak roughness measure, let us recall the notion of co-entropy [2, 15, 16, 18, 39]. Assume that $\pi = \{C_1, C_2, \ldots, C_m\} \in \Pi(U)$. Then the *co-entropy* of partition $\pi$ is defined as

$$E(\pi) = \frac{1}{|U|} \sum_{i=1}^{m} |C_i| \log |C_i|.$$

It has been known that the following identity holds

$$H(\pi) + E(\pi) = \log |U| \text{ for any } \pi \in \Pi(U),$$

where $H(\pi)$ is the entropy of partition $\pi$ [15, 16, 30, 34, 39] defined as

$$H(\pi) = - \sum_{i=1}^{m} \frac{|C_i|}{|U|} \log \frac{|C_i|}{|U|}.$$

Notice that a standard result of information theory assures the strict anti-monotonicity of entropy (see, for example, [30]):

$$\text{if } \pi \prec \sigma, \text{ then } H(\sigma) < H(\pi).$$

Therefore, we have the following strict monotonicity of co-entropy with respect to the partition ordering (a direct proof of this result can be found in [16]):

$$\text{if } \pi \prec \sigma, \text{ then } E(\pi) < E(\sigma).$$

This result has been proven in a roughness monotonicity theorem of [34], which is based on a lemma of the same paper. We thus find that $E$ is a partition measure on $U$. Noting that $E(\check{\pi}) = \log |U|$, we get the following proposition.

**Proposition 11.** *Define*

$$\beta_E(\pi, A) = \beta_P(\pi, A) \cdot \frac{E(\pi)}{\log |U|}$$

*for any $\pi \in \Pi(U)$ and $A \subseteq U$. Then $\beta_E$ is a strong Pawlak roughness measure.*

Let us remark that $\beta_P(\pi, A) \cdot E(\pi)$ was defined as the rough entropy of $A$ in [2]. Therefore, we may view $\beta_E(\pi, A)$ as a standardization of the rough entropy of $A$.

In [3], Bianucci et al. introduced a pseudo co-entropy related to a partition $\pi = \{C_1, C_2, \ldots, C_m\} \in \Pi(U)$ as follows:

$$E'(\pi) = \frac{1}{|U|} \sum_{i=1}^{m} |C_i|^2 \log |C_i|.$$

It has been proven in [3] that

$$\text{if } \pi \prec \sigma, \text{ then } E'(\pi) < E'(\sigma).$$

This property, together with the definition of $E'$, implies that $E'$ is also a partition measure on $U$. By using $E'(\check{\pi}) = |U| \log |U|$, we obtain the following proposition.

**Proposition 12.** *Define*

$$\beta_{E'}(\pi, A) = \beta_P(\pi, A) \cdot \frac{E'(\pi)}{|U| \log |U|}$$

*for any $\pi \in \Pi(U)$ and $A \subseteq U$. Then $\beta_{E'}$ is a strong Pawlak roughness measure.*

We end this section with one more strong Pawlak roughness measure arising from the concept of combination granulation introduced by Qian and Liang in [29].

Let $\pi = \{C_1, C_2, \ldots, C_m\} \in \Pi(U)$. Then the *combination granulation* of $\pi$, denoted by $CG(\pi)$, is defined as [29]

$$CG(\pi) = \sum_{i=1}^{m} \frac{|C_i|^2}{|U|^2} \cdot \frac{|C_i| - 1}{|U| - 1}.$$

It has been shown by Proposition 9 in [29] that

$$\text{if } \pi \prec \sigma, \text{ then } CG(\pi) < CG(\sigma).$$

This property, together with the definition of $CG$, implies that $CG$ is a partition measure on $U$. As $CG(\check{\pi}) = 1$, we obtain the following proposition.

**Proposition 13.** *Define*

$$\beta_{CG}(\pi, A) = \beta_P(\pi, A) \cdot CG(\pi)$$

*for any $\pi \in \Pi(U)$ and $A \subseteq U$. Then $\beta_{CG}$ is a strong Pawlak roughness measure.*

**Remark 2.** Thanks to the axiomatic approach, Propositions 11, 12, and 13 show us that the roughness measures $\beta_E$, $\beta_{E'}$, and $\beta_{CG}$ are of the properties stated in Corollary 3 and Propositions 5, 6, 7, and 8.

Let us calculate the above five strong Pawlak roughness measures for the sets and partitions in Example 1.

**Example 4.** *Let us revisit Example 1, where $U = \{a_1, a_2, a_3, a_4, a_5\}$, $\pi = \{\{a_1\}, \{a_2, a_3\}, \{a_4, a_5\}\}$, and $\sigma = \{\{a_1, a_2, a_3\}, \{a_4, a_5\}\}$. For $A = \{a_1, a_2, a_3, a_4\}$, we have already obtained that $\beta_P(\pi, A) = \beta_P(\rho, A) = 0.4$ in Example 1.*

*By a direct computation, we can readily get the following results:*

$$\begin{aligned}
\beta_X(\pi, A) = 0.102 &< \beta_X(\sigma, A) = 0.219; \\
\beta_L(\pi, A) = 0.144 &< \beta_L(\sigma, A) = 0.208; \\
\beta_E(\pi, A) = 0.138 &< \beta_E(\sigma, A) = 0.233; \\
\beta_{E'}(\pi, A) = 0.055 &< \beta_{E'}(\sigma, A) = 0.126; \\
\beta_{CG}(\pi, A) = 0.032 &< \beta_{CG}(\sigma, A) = 0.088.
\end{aligned}$$

*All of these are consistent with the fact that $\pi \prec \sigma$.*

## 5. Conclusion

In this paper, we have investigated roughness measure in an axiomatic way. The axiomatic definitions of roughness measure and partition measure have been provided. Based on this, we have given a unified construction of roughness measure, called strong Pawlak roughness measure, by combining partition measure into Pawlak's roughness measure. Some properties of the strong Pawlak roughness measure have been explored as well. In addition, we have shown that the existing roughness measures in [35] and [17] are two special instances of our strong Pawlak roughness measure, which supports our axiomatic definitions. The advantage of axiomatic approach is that it can bring together some seemingly different notions. As a result, we may study the properties of roughness measure in Definition 2 and weak roughness measure in Definition 3 instead of some specific measure.

As mentioned earlier, our axiomatic definition of roughness measure is largely dependent on partition measure, while the latter is based upon the cardinalities of all equivalence classes in a partition. Because partition measures can be defined from different perspectives, more axiomatic definitions and properties of roughness measure remain to be investigated. In addition, the present work focuses on the classical rough sets based on partitions. It would be interesting to examine our axiomatic approach in the framework of covering rough sets [6, 28, 43] or fuzzy rough sets [8]. Finally, in the face of so many roughness measures, the criterion for choice remains yet to be addressed.

## Acknowledgements

This work was supported by the National Natural Science Foundation of China under Grants 60821001, 60873191, and 60903152.